\documentclass[11pt,a4paper]{article}
\usepackage[hyperref]{acl2018}
\usepackage{times}
\usepackage{latexsym}
\usepackage{microtype}
\usepackage{graphicx}
\usepackage{subfigure}
\usepackage{algorithm}
\usepackage{algorithmic}
\usepackage{booktabs} 
\usepackage{Definitions}
\graphicspath{{./Figs/}}
\usepackage{url}

\aclfinalcopy 
\def\aclpaperid{1519} 

\title{LinkNBed: Multi-Graph Representation Learning with Entity Linkage }
\author{Rakshit Trivedi \Thanks{Correspondence: {\tt rstrivedi@gatech.edu}. Work done when the author interned at Amazon.} \\ College of Computing \\ Georgia Tech \\\\ {\bf Christos Faloutsos} \\ SCS, CMU \\ and Amazon.com \\\And
   Bunyamin Sisman \\ Amazon.com \\\\\\ {\bf Hongyuan Zha} \\  College of Computing \\ Georgia Tech \\\And
   Jun Ma \\ Amazon.com \\\\\\ {\bf Xin Luna Dong} \\ Amazon.com }

\date{}

\begin{document}
\maketitle
\begin{abstract}
Knowledge graphs have emerged as an important model for studying complex multi-relational data. This has given rise to the construction of numerous large scale but incomplete knowledge graphs encoding information extracted from various resources. An effective and scalable approach to jointly learn over multiple graphs and eventually construct a unified graph is a crucial next step for the success of knowledge-based inference for many downstream applications. To this end, we propose \textbf{LinkNBed}, a deep relational learning framework that learns entity and relationship representations across multiple graphs. We identify \textbf{entity linkage} across graphs as a vital component to achieve our goal.
We design a novel objective that leverage entity linkage and build an efficient multi-task training procedure. Experiments on link prediction and entity linkage demonstrate substantial improvements over the state-of-the-art relational learning approaches.

\end{abstract}

\section{Introduction}

\label{intro}
Reasoning over multi-relational data is a key concept in Artificial Intelligence and knowledge graphs have appeared at the forefront as an effective tool to model such multi-relational data. Knowledge graphs have found increasing importance due to its wider range of important applications such as information retrieval~\citep{DalDieAll14}, natural language processing~\citep{GabMar09}, recommender systems~\cite{CatCoh16}, question-answering~\cite{CuiXiaWanSonHwaWan17} and many more. This has led to the increased efforts in constructing numerous large-scale Knowledge Bases (e.g. Freebase~\citep{BolEvaParStuTay08}, DBpedia~\citep{AueBizKobLehCygIve07}, Google's Knowledge graph~\citep{DonGabHeiHorLaoMurStrSunZha14}, Yago~\citep{SucKasWei07} and NELL~\citep{CarBetKisSetHruMit10}), that can cater to these applications, by representing information available on the web in relational format.

All knowledge graphs share common drawback of \textit{incompleteness} and \textit{sparsity} and hence most existing relational learning techniques focus on using observed triplets in an incomplete graph to infer unobserved triplets for that graph~\citep{NicMurTreGab16}. Neural embedding techniques that learn vector space representations of entities and relationships have achieved remarkable success in this task. However, these techniques only focus on learning from a single graph. In addition to incompleteness property, these knowledge graphs also share a set of  overlapping entities and relationships with varying information about them. This makes a compelling case to design a technique that can learn over multiple graphs and eventually aid in constructing a unified giant graph out of them. While research on learning representations over single graph has progressed rapidly in recent years~\citep{NicTreKri11, DonGabHeiHorLaoMurStrSunZha14, TroWelRieGauBou16, BorUsuGarWesetal13, XiaHuaZhu16, YanYihHeGaoDen15}, there is a conspicuous lack of principled approach to tackle the unique challenges involved in learning across multiple graphs. 

One approach to multi-graph representation learning could be to first solve graph alignment problem to merge the graphs and then use existing relational learning methods on merged graph. Unfortunately, graph alignment is an important but still unsolved problem and there exist several techniques addressing its challenges ~\citep{LiuYan16, PerYakCha15, KouTonLub13, BunSta16} in limited settings. 
The key challenges for the graph alignment problem emanate from the fact that the real world data are noisy and intricate in nature. The noisy or sparse data make it difficult to learn robust alignment features, and data abundance leads to computational challenges due to the combinatorial permutations needed for alignment. These challenges are compounded in multi-relational settings due to heterogeneous nodes and edges in such graphs.

Recently, deep learning has shown significant impact in learning useful information over noisy, large-scale and heterogeneous graph data~\citep{RosZhoAhm17}. We, therefore, posit that combining graph alignment task with deep representation learning across multi-relational graphs has potential to induce a synergistic effect on both tasks. 
Specifically, we identify that a key component of graph alignment process---entity linkage---also plays a vital role in learning across graphs.
For instance, the embeddings learned over two knowledge graphs for an actor should be closer to one another compared to the embeddings of all the other entities. Similarly, the entities that are already aligned together across the two graphs should produce better embeddings due to the shared context and data. To model this phenomenon, we propose \textbf{LinkNBed}, a novel deep learning framework that jointly performs representation learning and graph linkage task.
To achieve this, we identify key challenges involved in the learning process and make the following contributions to address them:   
\begin{itemize}
\item We propose novel and principled approach towards jointly learning entity representations and entity linkage. The novelty of our framework stems from its ability to support linkage task across heterogeneous types of entities.
\item We devise a graph-independent inductive framework that learns functions to capture contextual information for entities and relations. It combines the structural and semantic information in individual graphs for joint inference in a principled manner.
\item Labeled instances (specifically positive instances for linkage task) are typically very sparse and hence we design a novel multi-task loss function where  entity linkage task is tackled in robust manner across various learning scenarios such as learning only with unlabeled instances or only with negative instances.
\item We design an efficient training procedure to perform joint training in linear time in the number of triples. We  demonstrate superior performance of our method on two datasets curated from Freebase and IMDB against state-of-the-art neural embedding methods.
\end{itemize}
 
\section{Preliminaries}
\label{bkgrnd}

\subsection{Knowledge Graph Representation}
A knowledge graph $\mathcal{G}$ comprises of set of facts represented as triplets ($e^s,r,e^o$) denoting the relationship $r$ between subject entity $e^s$  and object entity $e^o$. Associated to this knowledge graph, we have a set of attributes that describe observed characteristics of an entity. Attributes are represented as set of key-value pairs for each entity and an attribute can have null (missing) value for an entity. We follow \emph{Open World Assumption - triplets not observed in knowledge graph are considered to be missing but not false.} We assume that there are no duplicate triplets or self-loops.

\subsection{Multi-Graph Relational Learning} 

{\bf Definition.} Given a collection of knowledge graphs $\mathcal{G}$,  Multi-Graph Relational Learning refers to the the task of learning information rich representations of entities and relationships across graphs. The learned embeddings can further be used to infer new knowledge in the form of link prediction or learn new labels in the form of entity linkage. We motivate our work with the setting of two knowledge graphs where given two graphs $G_1, G_2 \in \mathcal{G}$, the task is to match an entity $e_{G_1} \in G_1$ to an entity $e_{G_2} \in G_2$ if they represent the same real-world entity. We discuss a straightforward extension of this 
\vspace{0.2cm}
setting to more than two graphs in Section 7. 
{\bf Notations.} Let $X$ and $Y$ represent realization of two such knowledge graphs extracted from two different sources. Let $n_e^X$ and $n_e^Y$ represent number of entities in $X$ and $Y$ respectively. Similarly, $n_r^X$ and $n_r^Y$ represent number of relations in $X$ and $Y$. We combine triplets from both $X$ and $Y$ to obtain set of all observed triplets $\mathcal{D} = \{(e^s, r, e^o)_p\}_{p=1}^{P}$ where $P$ is total number of available records across from both graphs. Let $\mathcal{E}$ and $\mathcal{R}$ be the set of all entities and all relations in $\mathcal{D}$ respectively. Let $|\mathcal{E}| = n$ and $|\mathcal{R}| = m$. In addition to $\mathcal{D}$, we also have set of linkage labels $\mathcal{L}$ for entities between $X$ and $Y$. Each record in $\mathcal{L}$ is represented as triplet ($e^X \in X$, $e^Y \in Y$, $l \in \{0,1\}$) where $l=1$ when the entities are matched and $l=0$ otherwise.  
\section{Proposed Method: LinkNBed }
\label{model}

We present a novel \textit{inductive multi-graph} relational learning framework that learns a set of aggregator functions capable of ingesting various contextual information for both entities and relationships in multi-relational graph. These functions encode the ingested structural and semantic information into low-dimensional entity and relation embeddings. Further, we use these representations to learn a relational score function that computes how two entities are likely to be connected in a particular relationship. The key idea behind this formulation is that when a triplet is observed, the relationship between the two entities can be explained using various contextual information such as local \textit{neighborhood} features of both entities, \textit{attribute} features of both entities and \textit{type} information of the entities which participate in that relationship. 

We outline two key insights for establishing the relationships between embeddings of the entities over multiple graphs in our framework:\\
\textit{\textbf{Insight 1 (Embedding Similarity)}}: If the two entities $e^X \in X$ and $e^Y \in Y$ represent the same real-world entity then their embeddings $\mathbf{e^X}$ and $\mathbf{e^Y}$ will be close to each other. \\
\textit{\textbf{Insight 2 (Semantic Replacement)}}: For a given triplet $t = (e^s, r, e^o) \in X$, denote $g(t)$ as the function that computes a relational score for $t$ using entity and relation embeddings. If there exists a matching entity $e^{s'} \in Y$ for $e^s \in X$, denote $t' = (e^{s'},r,e^o)$ obtained after replacing $e^s$ with $e^{s'}$. In this case, $g(t) \sim g(t')$ i.e. score of triplets $t$ and $t'$ will be similar. 

\begin{figure*}[t]
\includegraphics[width = 1\textwidth, height = 0.35\textwidth]{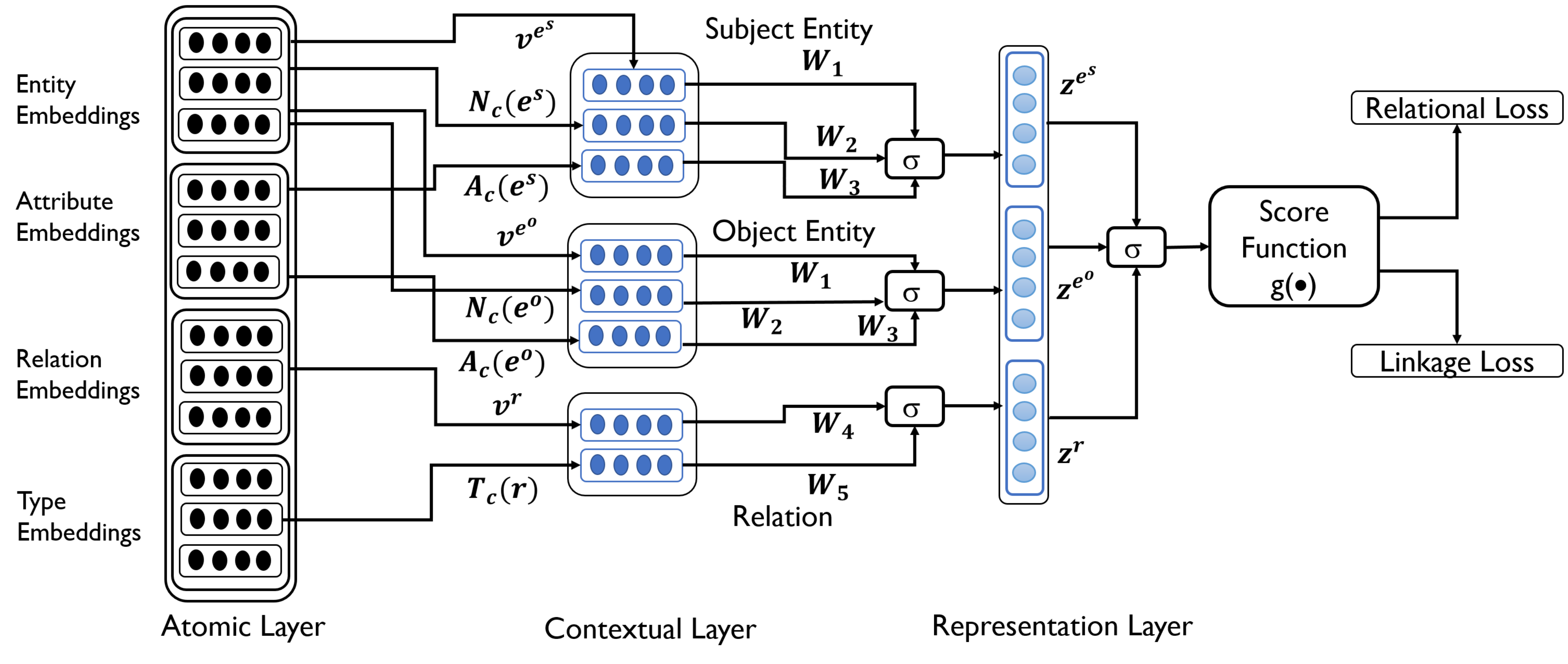}
\caption{LinkNBed Architecture Overview - one step score computation for a given triplet ($e^s, r, e^o$). The Attribute embeddings are not simple lookups but they are learned as shown in Eq~\ref{eq:att_embed}}
\label{fig:deepGL}
\end{figure*}
For a triplet $(e^s, r , e^o) \in \mathcal{D}$, we describe encoding mechanism of LinkNBed as three-layered architecture that computes the final output representations of $\mathbf{z}^{r}, \mathbf{z}^{e^s}, \mathbf{z}^{e^o}$ for the given triplet. Figure~\ref{fig:deepGL} provides an overview of LinkNBed architecture and we describe the three steps below:

\subsection{Atomic Layer}
Entities, Relations, Types and Attributes are first encoded in its basic vector representations. We use these basic representations to derive more complex contextual embeddings further.\\
{\bf Entities, Relations and Types.} The embedding vectors corresponding to these three components are learned as follows:
\begin{align}
\label{eq:ent_embed}
	\mathbf{v^{e^s}} &= f(\mathbf{W^E} \mathbf{e^s} ) &   \mathbf{v^{e^o}} &= f(\mathbf{W^E} \mathbf{e^o} ) \\
	\mathbf{v^{r}} &= f(\mathbf{W^R} \mathbf{r} ) & \mathbf{v^{t}} &= f(\mathbf{W^T} \mathbf{t} )
\end{align}
where $\mathbf{v^{e^s}}$,$\mathbf{v^{e^o}} \in \mathbb{R}^{d}$. $\mathbf{e^s}$, $\mathbf{e^o} \in \mathbb{R}^{n}$ are ``one-hot" representations of $e^s$ and $e^o$ respectively. $\mathbf{v^{r}} \in \mathbb{R}^{k}$ and $\mathbf{r} \in \mathbb{R}^{m}$ is ``one-hot" representation 
of $r$. $\mathbf{v^{t}} \in \mathbb{R}^{q}$ and $\mathbf{t} \in \mathbb{R}^{z}$ is "one-hot" representation of $t$ . $\mathbf{W^E} \in \mathbb{R}^{d \times n}$, $\mathbf{W^R} \in \mathbb{R}^{k \times m}$ and $\mathbf{W^T} \in \mathbb{R}^{q \times z}$ are the entity, relation and type embedding matrices respectively. $f$ is a nonlinear activation function (Relu in our case). $\mathbf{W^E}$, $\mathbf{W^R}$ and $\mathbf{W^T}$ can be initialized randomly or using pre-trained word embeddings or vector compositions based on name phrases of components~\cite{SocCheManNg13}.\\
{\bf Attributes.} For a given attribute $a$ represented as key-value pair, we use paragraph2vec ~\cite{LeMik14} type of embedding network to learn attribute embedding. Specifically, we represent attribute embedding vector as:
\begin{align}
\label{eq:att_embed}
	\mathbf{a} &= f(\mathbf{W^{key}} \mathbf{a_{key}} + \mathbf{W^{val}} \mathbf{a_{val}} )
\end{align}

where $\mathbf{a} \in \mathbb{R}^{y}$, $\mathbf{a_{key}} \in \mathbb{R}^{u}$ and $\mathbf{a_{val}} \in \mathbb{R}^{v}$.
$\mathbf{W^{key}} \in \mathbb{R}^{y \times u}$ and $\mathbf{W^{val}} \in \mathbb{R}^{y \times v}$. $\mathbf{a_{key}}$
will be ``one-hot" vector and $\mathbf{a_{val}}$ will be feature vector.
Note that the dimensions of the embedding vectors do not necessarily need to be the same.

\subsection{Contextual Layer}

While the entity and relationship embeddings described above help to capture very generic latent features, embeddings can be further enriched to capture structural information,
attribute information and type information to better explain the existence of a fact. Such information can be modeled as context of nodes and edges in the graph. To this end, we design the following canonical aggregator function that learns  various contextual information by aggregating over relevant embedding vectors:
\begin{align}
\label{eq:context}
	\mathbf{c}(z) &= \textrm{AGG}(\{\mathbf{z'}, \forall z' \in C(z)\})
\end{align}
where $\mathbf{c}(z)$ is the vector representation of the aggregated contextual information for component $z$. Here, component $z$ can be either an entity or a relation. $C(z)$ is the set of components in the context of $z$ and $\mathbf{z'}$ correspond to the vector embeddings of those components. \textrm{AGG} is the aggregator function which can take many forms such Mean, Max, Pooling or more complex LSTM based aggregators. It is plausible that different components in a context may have varied impact on the component for which the embedding is being learned. To account for this, we employ a soft attention mechanism where we learn attention coefficients to weight components based on their impact before  aggregating them. We modify Eq.~\ref{eq:context} as:
\vspace{-0.1cm}
\begin{align}
\label{eq:attn1}
	\mathbf{c}(z) &= \textrm{AGG}(\mathbf{q}(z) * \{\mathbf{z'}, \forall z' \in C(z)\})
\end{align}
where
\begin{align}
\label{eq:attn2}
	\mathbf{q}(z) &= \frac{\exp(\theta_z)}{\sum\limits_{z' \in C(z)} \exp(\theta_{z'}) } 
\end{align}
and $\theta_z$'s are the parameters of attention model.\\
Following contextual information is modeled in our framework:\\\\
{\bf Entity Neighborhood Context $\mathbf{N_c}(e) \in \mathbb{R}^d$.} Given a triplet $(e^s,r,e^o)$, the neighborhood context for an entity $e^s$ will be the nodes located near $e^s$ other than the node $e^o$. This will capture the effect of local neighborhood in the graph surrounding $e^s$ that drives $e^s$ to participate in fact $(e^s,r,e^o)$. We use Mean as aggregator function. As there can be large number of neighbors, we collect the neighborhood set for each entity as a pre-processing step using a random walk method. Specifically, given a node $e$, we run $k$ rounds of random-walks of length $l$ following~\cite{HamYinLes17} and create set $\mathcal{N}(e)$ by adding all unique nodes visited across these walks. This context can be similarly computed for object entity.\\\\
{\bf Entity Attribute Context $\mathbf{A_c}(e) \in \mathbb{R}^y$.} For an entity $e$, we collect all attribute embeddings for $e$ obtained from Atomic Layer and learn aggregated information over them using Max operator given in Eq.~\ref{eq:context}.\\
{\bf Relation Type Context $\mathbf{T_c}(r) \in \mathbb{R}^q$.} We use type context for relation embedding i.e. for a given relationship $r$, this context aims at capturing the effect of type of entities that have participated in this relationship. For a given triplet $(e^s, r , e^o)$, type context for relationship $r$ is computed by  aggregation with mean over type embeddings corresponding to the context of $r$. Appendix C provides specific forms of contextual information.
\vspace{-0.2cm}
\subsection{Representation Layer}
Having computed the atomic and contextual embeddings for a triplet $(e^s, r, e^o)$, we obtain the final embedded representations of entities and relation in the triplet using the following formulation:

\begin{align}
\label{eq:repr}
\begin{split}
	\mathbf{z^{e^s}} &= \sigma(\underbrace{\mathbf{W_1v^{e^s}}}_\text{Subject Entity Embedding} + \underbrace{\mathbf{W_2 N_c}(e^s)}_\text{Neighborhood Context}\\  &+ \underbrace{\mathbf{W_3 A_c}(e^s))}_\text{Subject Entity Attributes}
	\end{split} \\
	\begin{split}
	\mathbf{z^{e^o}} &= \sigma(\underbrace{\mathbf{W_1v^{e^o}}}_\text{Object Entity Embedding} + \underbrace{\mathbf{W_2 N_c}(e^o)}_\text{Neighborhood Context}\\ &+ \underbrace{\mathbf{W_3 A_c}(e^o))}_\text{Object Entity Attributes}
	\end{split} \\
	\mathbf{z^{r}} &= \sigma(\underbrace{\mathbf{W_4v^{r}}}_\text{Relation Embedding} + \underbrace{\mathbf{W_5 T_c}(r))}_\text{Entity Type Context}
\end{align}\\
where $\mathbf{W_1}, \mathbf{W_2} \in \mathbb{R}^{d \times d}$, $\mathbf{W_3} \in \mathbb{R}^{d \times y}$,
$\mathbf{W_4} \in \mathbb{R}^{d \times k}$ and $\mathbf{W_5} \in \mathbb{R}^{d \times q}$. $\sigma$ is nonlinear activation function -- generally Tanh or Relu. \\
Following is the rationale for our formulation: An entity's representation can be enriched by encoding information about the local neighborhood features and attribute information associated with the entity in addition to its own latent features. Parameters $\mathbf{W_1}, \mathbf{W_2}, \mathbf{W_3}$ learn to capture these different aspects and map them into the entity embedding space. Similarly, a relation's representation can be enriched by encoding information about entity types that participate in that relationship in addition to its own latent features. Parameters $\mathbf{W_4}, \mathbf{W_5}$ learn to capture these  aspects and map them into the relation embedding space. Further, as the ultimate goal is to jointly learn over multiple graphs, shared parameterization in our model facilitate the propagation of information across graphs thereby making it a graph-independent inductive model. The flexibility of the model stems from the ability to shrink it (to a very simple model considering atomic entity and relation embeddings only) or expand it (to a complex model by adding different contextual information) without affecting any other step in the learning procedure.

\subsection{Relational Score Function}

Having observed a triplet $(e^s,r, e^o)$, we first use Eq. 7, 8 and 9 to compute entity and relation representations. We then use these embeddings to capture relational interaction between two entities using the following score function $g(\cdot)$:
\begin{align}
\label{eq:score}
	g(e^s, r, e^o) &= \sigma(\mathbf{z}^{r^T} \cdot(\mathbf{z}^{e^s} \odot \mathbf{z}^{e^o}))
\end{align} 
where $\mathbf{z}^{r}, \mathbf{z}^{e^s}, \mathbf{z}^{e^o} \in \mathbb{R}^d$ are $d$-dimensional representations of entity and relationships as described below. $\sigma$ is the nonlinear activation function and $\odot$ represent element-wise product.

\section{Efficient Learning Procedure}

\subsection{Objective Function}
The complete parameter space of the model can be given by: $\mathbf{\Omega = \{\{W_i\}_{i=1}^5, W^E, W^R, W^{key}, W^{val}, W^t ,\Theta\}}$. To learn these parameters, we design a novel multi-task objective function that jointly trains over two graphs.
As identified earlier, the goal of our model is to leverage the available linkage information across graphs for optimizing the entity and relation embeddings such that they can explain the observed triplets across the graphs. Further, we want to leverage these optimized embeddings to match entities across graphs and expand the available linkage information. To achieve this goal, we define following two different loss functions catering to each learning task and jointly optimize over them as a multi-task objective to learn model parameters:\\\\ 
{\bf Relational Learning Loss.} This is conventional loss function used to learn knowledge graph embeddings. Specifically, given a p-th triplet $(e^s, r, e^o)_p$ from training set $\mathcal{D}$, we sample $C$ negative samples by replacing either head or tail entity and define a contrastive max margin function as shown in ~\cite{SocCheManNg13}:
\begin{equation}
\label{eq:gen}
\begin{split}
	L_{rel} &=  \sum\limits_{c=1}^{C} \max(0, \gamma - g(e^s_p,r_p,e^o_p) \\ &+ g'(e^s_c,r_p,e^o_p))
\end{split} 
\end{equation}
where, $\gamma$ is margin, $e^s_c$ represent corrupted entity and $g'(e^s_c,r_p,e^o_p)$ represent corrupted triplet score.\\\\
{\bf Linkage Learning Loss:} We design a novel loss function to leverage pairwise label set $\mathcal{L}$. Given a triplet $(e^s_X, r_X, e^o_X)$ from knowledge graph $X$, we first find the entity $e_Y^+$ from graph $Y$ that represent the same real-world entity as $e^s_X$. We then replace $e^s_X$ with $e_Y^+$ and compute score $g(e_Y^+,r_X,e^o_X)$. Next, we find set of all entities $E_Y^-$ from graph $Y$ that has a negative label with entity $e^s_X$. We consider them analogous to the negative samples we generated for Eq.~\ref{eq:gen}. We then propose the label learning loss function as:
\begin{equation}
\label{eq:res}
\begin{split}
	L_{lab} &=  \sum\limits_{z=1}^{Z} \max(0, \gamma - g(e_Y^+,r_X,e^o_X) \\ &+ (g'(e_Y^-,r_X,e^o_X)_z))
\end{split} 
\end{equation}
where, $Z$ is the total number of negative labels for $e_X$. $\gamma$ is margin which is usually set to 1 and $e_Y^- \in E_Y^-$ represent entity from graph $Y$ with which entity $e^s_X$ had a negative label. Please note that this applies symmetrically for the triplets that originate from graph $Y$ in the overall dataset. Note that if both entities of a triplet have labels, we will include both cases when computing the loss. Eq.~\ref{eq:res} is inspired by \textit{Insight 1} and \textit{Insight 2} defined earlier in Section 2.
Given a set $\mathcal{D}$ of $N$ observed triplets across two graphs, we define complete multi-task objective as:
\begin{equation}
\label{eq:fin}
	\mathbf{L}(\mathbf{\Omega}) =  \sum\limits_{i=1}^{N} [b \cdot L_{rel} + (1-b) \cdot L_{lab}] +  \lambda\left\lVert \mathbf{\Omega} \right\rVert_2^2
\end{equation}
where $\mathbf{\Omega}$ is set of all model parameters and $\lambda$ is regularization hyper-parameter. $b$ is weight hyper-parameter used to attribute importance to each task. We train with mini-batch SGD procedure (Algorithm~\ref{alg:alg1}) using Adam Optimizer.
\begin{algorithm}[t!]
   \caption{LinkNBed mini-batch Training}
   \label{alg:alg1}
\begin{algorithmic}
   \STATE {\bfseries Input:} Mini-batch $\mathcal{M}$, Negative Sample Size $C$, Negative Label Size $Z$, Attribute data $att\_data$, Neighborhood data $nhbr\_data$, Type data $type\_data$, Positive Label Dict $pos\_dict$, Negative Label Dict $neg\_dict$
   \STATE {\bfseries Output:} Mini-batch Loss $\mathcal{L_M}$.
   \STATE $\mathcal{L_M} = 0$
   \STATE score\_pos = []; score\_neg = []; score\_pos\_lab = []; score\_neg\_lab = [] 
   \FOR{$i=0$ {\bfseries to} size($\mathcal{M}$)}
   \STATE input\_tuple = $\mathcal{M}[i]$ = ($e^s, r, e^o$)
   \STATE sc = compute\_triplet\_score($e^s, r, e^o$) (Eq.~\ref{eq:score})
   \STATE score\_pos.append(sc)
   \FOR{$j=0$ {\bfseries to} $C$}
   \STATE Select $e^s_c$ from entity list such that   $e^s_c \ne e^s$ and $e^s_c \ne e^o$ and $(e^s_c, r, e^o) \notin \mathcal{D}$ 
   \STATE sc\_neg = compute\_triplet\_score($e^s_c, r, e^o$)
   \STATE score\_neg.append(sc\_neg)
   \ENDFOR
   \IF {$e^s$ in $pos\_dict$}
   \STATE $e^{s+}$ = positive label for $e^s$ 
   \STATE sc\_pos\_l = compute\_triplet\_score($e^{s+}, r, e^o$) 
   \STATE score\_pos\_lab.append(sc\_pos\_l)
   \ENDIF
   \FOR{$k=0$ {\bfseries to} $Z$}
   \STATE Select $e^{s-}$ from $neg\_dict$
   \STATE sc\_neg\_l = compute\_triplet\_score($e^{s-}, r, e^o$)
   \STATE score\_neg\_lab.append(sc\_neg\_l)
   \ENDFOR
   \ENDFOR
   \STATE $\mathcal{L_M}\mathrel{+}=$ compute\_minibatch\_loss(score\_pos, score\_neg, score\_pos\_lab, score\_neg\_lab) (Eq.~\ref{eq:fin})
   \STATE Back-propagate errors and update parameters $\mathbf{\Omega}$
   \STATE {\bf return} $\mathcal{L_M}$
\end{algorithmic}
\end{algorithm}\\
{\bf Missing Positive Labels.} It is expensive to obtain positive labels across multiple graphs and hence it is highly likely that many entities will not have positive labels available. For those entities, we will modify Eq.~\ref{eq:res} to use the original triplet $(e^s_X, r_X, e^o_X)$ in place of perturbed triplet $g(e_Y^+,r_X,e^o_X)$ for the positive label. The rationale here again arises from \textit{Insight 2} wherein embeddings of two duplicate entities should be able to replace each other without affecting the score. \\
{\bf Training Time Complexity.} Most contextual information is pre-computed and available to all training steps which leads to constant time embedding lookup for those context. But for attribute network, embedding needs to be computed for each attribute separately and hence the complexity to compute score for one triplet is $\mathcal{O}(2a)$ where $a$ is number of attributes. Also for training, we generate $C$ negative samples for relational loss function and use $Z$ negative labels for label loss function. Let $k = C + Z$. Hence, the training time complexity for a set of $n$ triplets will be $\mathcal{O}(2ak*n)$ which is linear in number of triplets with a constant factor as $ak << n$ for real world knowledge graphs. This is desirable as the number of triplets tend to be very large per graph in multi-relational settings.\\
{\bf Memory Complexity.} We borrow notations from~\citep{NicMurTreGab16} and describe the parameter complexity of our model in terms of the number of each component and corresponding embedding dimension requirements. Let $H_a = 2*N_eH_e + N_rH_r + N_tH_t + N_kH_k + N_vH_v$. The parameter complexity of our model is: $H_a * (H_b + 1)$. Here, $N_e$, $N_r$, $N_t$, $N_k$, $N_v$ signify number of entities, relations, types, attribute keys and vocab size of attribute values across both datasets. Here $H_b$ is the output dimension of the hidden layer.
\section{Experiments}
\subsection{Datasets}
We evaluate LinkNBed and baselines on two real world knowledge graphs: D-IMDB (derived from large scale IMDB data snapshot) and D-FB (derived from large scale Freebase data snapshot). Table~\ref{tab:data} provides statistics for our final dataset used in the experiments. 
Appendix B.1 provides complete details about dataset processing.

\begin{table}[ht!]
\resizebox{0.5\textwidth}{!}{
\centering
\begin{tabular}{cccccc}
\toprule
Dataset & \# Entities & \# Relations & \# Attributes & \# Entity & \# Available\\
Name &&&&Types&Triples\\
\midrule
D-IMDB & 378207 & 38 & 23 & 41 & 143928582 \\
D-FB & 39667 & 146 & 69 & 324 & 22140475 \\
\bottomrule
\label{tab:data}
\end{tabular}
}
\vspace{-0.2cm}
\caption{Statistics for Datasets: D-IMDB and D-FB}
\end{table}
\vspace{-0.2cm}
\subsection{Baselines}
We compare the performance of our method against state-of-the-art representation learning baselines that use neural embedding techniques to learn entity and relation representation. Specifically, we consider compositional methods of RESCAL~\citep{NicTreKri11} as basic matrix factorization method, DISTMULT~\citep{YanYihHeGaoDen15} as simple multiplicative model good for capturing symmetric relationships, and Complex~\citep{TroWelRieGauBou16}, an upgrade over DISTMULT that can capture asymmetric relationships using complex valued embeddings. We also compare against translational model of STransE that combined original structured embedding with TransE and has shown state-of-art performance in benchmark testing~\citep{KadBajKle17}. Finally, we compare with GAKE~\citep{FenHuaYanZhu16}, a model that captures context in entity and relationship representations.\\\\
In addition to the above state-of-art models, we analyze the effectiveness of different components of our model by comparing with various versions that use partial information. Specifically, we report results on following variants:\\{\bf LinkNBed - Embed Only.} Only use entity embeddings, {\bf LinkNBed - Attr Only.} Only use Attribute Context, {\bf LinkNBed - Nhbr Only.} Only use Neighborhood Context, {\bf LinkNBed - Embed + Attr.} Use both Entity embeddings and Attribute Context, {\bf LinkNBed - Embed + Nhbr.} Use both Entity embeddings and Neighbor Context and {\bf LinkNBed - Embed All.} Use all three Contexts.
\vspace{-0.1cm}
\subsection{Evaluation Scheme}

We evaluate our model using two inference tasks:\\
{\bf Link Prediction.} Given a test triplet $(e^s, r, e^o)$, we first score this triplet using Eq.~\ref{eq:score}. We then replace $e^o$ with all other entities in the dataset and filter the resulting set of triplets as shown in \cite{BorUsuGarWesetal13}. We score the remaining set of perturbed triplets using Eq.~\ref{eq:score}. All the scored triplets are sorted based on the scores and then the rank of the ground truth triplet is used for the evaluation. We use this ranking mechanism to compute HITS@10 (predicted rank $\leq$ 10) and reciprocal rank ($\frac{1}{rank}$) of each test triplet. We report the mean over all test samples. \\\\
{\bf Entity Linkage.} In alignment with \textit{Insight 2}, we pose a novel evaluation scheme to perform entity linkage. Let there be two ground truth test sample triplets: $(e_X, e_Y^+, 1)$ representing a positive duplicate label and $(e_X, e_Y^-, 0)$ representing a negative duplicate label. Algorithm~\ref{alg:alg2} outlines the procedure to compute linkage probability or score $q$ ($ \in [0,1]$) for the pair $(e_X, e_Y)$. We use $L1$ distance between the two vectors  analogous to Mean Absolute Error (MAE). 
In lieu of hard-labeling test pairs, we use score $q$ to compute Area Under the Precision-Recall Curve (AUPRC).

\begin{algorithm}[t!]
   \caption{Entity Linkage Score Computation}
   \label{alg:alg2}
\begin{algorithmic}
   \STATE {\bfseries Input:} Test pair -- $(e_X \in X, e_Y \in Y)$.
   \STATE {\bfseries Output:} Linkage Score -- $q$.
   \STATE 
   \STATE {\bf 1.} Collect all triplets involving $e_X$ from graph $X$ and all triplets involving $e_Y$ from graph $Y$ into a combined set $\mathcal{O}$. Let $|\mathcal{O}| = k$.
   \STATE {\bf 2.} Construct $S_{orig} \in \mathbb{R}^k$. 
   \STATE For each triplet $o \in \mathcal{O}$, compute score $g(o)$ using Eq.~\ref{eq:score} and store the score in $S_{orig}$.
   \STATE {\bf 3.} Create triplet set $\mathcal{O}'$ as following:
   \IF{$o \in \mathcal{O}$ contain $e^X \in X$} 
   \STATE Replace $e^X$ with $e^Y$ to create perturbed triplet $o'$ and store it in $\mathcal{O}'$
   \ENDIF
	\IF{$o \in \mathcal{O}$ contain $e^Y \in Y$} 
   \STATE Replace $e^Y$ with $e^X$ to create perturbed triplet $o'$ and store it in $\mathcal{O}'$
   \ENDIF   
   \STATE {\bf 4.} Construct $S_{repl} \in \mathbb{R}^k$. 
   \STATE For each triplet $o' \in \mathcal{O'}$, compute score $g(o')$ using Eq.~\ref{eq:score} and store the score in $S_{repl}$.
   \STATE {\bf 5.} Compute $q$.
   \STATE Elements in $S_{orig}$ and $S_{repl}$ have one-one correspondence so take the mean absolute difference:
   \STATE $q$ = $|S_{orig}$ - $S_{repl}|_1$
   \STATE \textbf{return} $q$
\end{algorithmic}
\end{algorithm}

For the baselines and the unsupervised version (with no labels for entity linkage) of our model, we use second stage multilayer Neural Network as classifier for evaluating entity linkage. Appendix B.2 provides training configuration details.
\subsection{Predictive Analysis}

\textbf{Link Prediction Results.} We train LinkNBed model jointly across two knowledge graphs and then perform inference over individual graphs to report link prediction reports. For baselines, we train each baseline on individual graphs and use parameters specific to the graph to perform link prediction inference over each individual graph. Table~\ref{tab:link_pred} shows link prediction performance for all methods. Our model variant with attention mechanism outperforms all the baselines with $4.15\%$ improvement over single graph state-of-the-art Complex model on D-IMDB and $8.23\%$ improvement on D-FB dataset. D-FB is more challenging dataset to learn as it has a large set of sparse relationships, types and attributes and it has an order of magnitude lesser relational evidence (number of triplets) compared to D-IMDB. Hence, LinkNBed's pronounced improvement on D-FB demonstrates the effectiveness of the model. The simplest version of LinkNBed with only entity embeddings resembles DISTMULT model with different objective function. Hence closer performance of those two models aligns with expected outcome. We observed that the Neighborhood context alone provides only marginal improvements while the model benefits more from the use of attributes. Despite being marginal, attention mechanism also  improves accuracy for both datasets. 
Compared to the baselines which are obtained by trained and evaluated on individual graphs, our superior performance demonstrates the effectiveness of multi-graph learning.

\begin{table}[t]
\resizebox{0.5\textwidth}{!}{
\centering
\begin{tabular}{ccccc}
\toprule
Method & D-IMDB-HITS10  &  D-IMDB-MRR & D-FB-HITS10 &  D-FB-MRR \\
\midrule
RESCAL & 75.3 & 0.592 & 69.99 & 0.147  \\
DISTMULT & 79.5 & 0.691 & 72.34 & 0.556  \\
Complex & 83.2 & 0.725 & 75.67 & 0.629 \\
STransE & 80.7  & 0.421 & 69.87 & 0.397  \\
GAKE & 69.5  & 0.114 & 63.22 & 0.093  \\
\midrule
\midrule
LinkNBed-Embed Only & 79.9 & 0.612 & 73.2 & 0.519  \\
LinkNBed-Attr Only & 82.2 & 0.676 & 74.7 & 0.588  \\
LinkNBed-Nhbr Only & 80.1 & 0.577 & 73.4 & 0.572 \\
LinkNBed-Embed + Attr & 84.2 & 0.673 & 78.39 & 0.606  \\
LinkNBed-Embed + Nhbr & 81.7 & 0.544 & 73.45 & 0.563  \\
LinkNBed-Embed All & 84.3 & 0.725 & 80.2 & 0.632  \\
LinkNBed-Embed All (Attention) & {\bf 86.8} & {\bf 0.733} & {\bf 81.9} & {\bf 0.677}  \\
\midrule
\midrule
{\bf Improvement (\%)} & {\bf 4.15}  & {\bf 1.10} & {\bf 7.61} & {\bf 7.09}  \\
\bottomrule
\label{tab:link_pred}
\end{tabular}
}
\caption{Link Prediction Results on both datasets}
\vspace{-0.3cm}
\end{table}

\textbf{Entity Linkage Results.}
We report entity linkage results for our method in two settings: a.) Supervised case where we train using both the objective functions. b.) Unsupervised case where we learn with only the relational loss function. The latter case resembles the baseline training where each model is trained separately on two graphs in an unsupervised manner. For performing the entity linkage in unsupervised case for all models, we first train a second stage of simple neural network classifier and then perform inference. In the supervised case, we use Algorithm~\ref{alg:alg2} for performing the inference. Table~\ref{tab:linkage} demonstrates the performance of all methods on this task. Our method significantly outperforms all the baselines with $33.86\%$ over second best baseline in supervised case and $17.35\%$ better performance in unsupervised case. The difference in the performance of our method in two cases demonstrate that the two training objectives are helping one another by learning across the graphs.
GAKE's superior performance on this task compared to the other state-of-the-art relational baselines shows the importance of using contextual information for entity linkage. Performance of other variants of our model again demonstrate that attribute information is more helpful than neighborhood context and attention provides  marginal improvements. We provide further insights with examples and detailed discussion on  entity linkage task in Appendix A.

\begin{table}[t]
\resizebox{0.5\textwidth}{!}{
\centering
\begin{tabular}{ccc}
\toprule
Method &  AUPRC (Supervised) &  AUPRC (Unsupervised)\\
\midrule
RESCAL & - & 0.327  \\
DISTMULT & - & 0.292  \\
Complex & - & 0.359 \\
STransE & -  & 0.231  \\
GAKE & -  & {\bf 0.457} \\
\midrule
\midrule
LinkNBed-Embed Only & 0.376 & 0.304  \\
LinkNBed-Attr Only & 0.451 & 0.397 \\
LinkNBed-Nhbr Only & 0.388 & 0.322  \\
LinkNBed-Embed + Attr & 0.512 & 0.414  \\
LinkNBed-Embed + Nhbr & 0.429 & 0.356  \\
LinkNBed-Embed All & 0.686 & 0.512 \\
LinkNBed-Embed All (Attention) & {\bf 0.691} & {\bf 0.553}  \\
\midrule
\midrule
{\bf Improvement (\%)} & {\bf 33.86} & {\bf 17.35}\\
\bottomrule
\label{tab:linkage}
\end{tabular}
}
\caption{Entity Linkage Results - Unsupervised case uses classifier at second step}
\vspace{-0.3cm}
\end{table}
\section{Related Work}
\subsection{Neural Embedding Methods for Relational Learning}
{\bf Compositional Models} learn representations by various composition operators on entity and relational embeddings. These models are multiplicative in nature and highly expressive but often suffer from scalability issues. Initial models include RESCAL~\citep{NicTreKri11} that uses a relation specific weight matrix to explain triplets via pairwise interactions of latent features, Neural Tensor Network~\citep{SocCheManNg13}, more expressive model that combines a standard NN layer with a bilinear
tensor layer and ~\cite{DonGabHeiHorLaoMurStrSunZha14} that  employs a concatenation-projection
method to project entities and relations to lower dimensional space. Later, many sophisticated models (Neural Association Model~\citep{LiuJiaEvdLinZhuXiaWeiHu16}, HoLE~\citep{NicRosPog16}) have been proposed. Path based composition models \citep{TouLinYihPooetal16} and contextual models GAKE~\citep{FenHuaYanZhu16} have been recently studied to capture more information from graphs. Recently, model like Complex~\citep{TroWelRieGauBou16} and Analogy~\citep{LiuWuYan17} have demonstrated state-of-the art performance on relational learning tasks.
{\bf Translational Models} (~\citep{BorGloWesBen14}, ~\citep{BorWesColBen11},~\citep{BorUsuGarWesetal13},~\citep{WanZhaFenChe14},~\citep{LinLiuSunZhu15}, \citep{XiaHuaZhu16}) learn representation by employing translational operators on the embeddings and optimizing based on their score. They offer an additive and efficient alternative to expensive multiplicative models. Due to their simplicity, they often loose expressive power. For a comprehensive survey of relational learning methods and empirical comparisons, we refer the readers to ~\citep{NicMurTreGab16},~\citep{KadBajKle17},~\citep{TouChe15} and~\citep{YanYihHeGaoDen15}. None of these methods address multi-graph relational learning and cannot be adapted to tasks like entity linkage in straightforward manner.
\vspace{-0.2cm}
\subsection{Entity Resolution in Relational Data}
Entity Resolution refers to resolving entities available in knowledge graphs with entity mentions in text. ~\citep{DreMcNRaoGerFin10} proposed entity disambiguation method for KB population, ~\citep{HeLiuLiZhoZhaWan13} learns entity embeddings for resolution, ~\citep{HuaHecJi15} propose a sophisticated DNN architecture for resolution,~\citep{CamLiDagAceGeyCamPri16} proposes entity resolution across multiple social domains,~\citep{FanZhaWanCheLi16} jointly embeds text and knowledge graph to perform resolution while~\citep{GloLazChaSubRinPer16} proposes Attention Mechanism for Collective Entity Resolution.
\vspace{-0.2cm}
\subsection{Learning across multiple graphs}
Recently, learning over multiple graphs have gained traction. ~\citep{LiuYan16} divides a multi-relational graph into multiple homogeneous graphs and learns associations across them by employing product operator. Unlike our work, they do not learn across multiple multi-relational graphs. ~\citep{PujGet16} provides logic based insights for cross learning, ~\citep{PerYakCha15} does pairwise entity matching across multi-relational graphs and is very expensive,~\citep{CheTiaYanZan17} learns embeddings to support multi-lingual learning and Big-Align~\citep{KouTonLub13} tackles graph alignment problem efficiently for bipartite graphs. None of these methods learn latent representations or jointly train graph alignment and learning which is the goal of our work.

\section{Concluding Remarks and Future Work}

We present a novel relational learning framework that learns entity and relationship embeddings across multiple graphs. The proposed representation learning framework leverage an efficient learning and inference procedure which takes into account the duplicate entities representing the same real-world entity in a multi-graph setting. We demonstrate superior accuracies on link prediction and entity linkage tasks compared to the existing approaches that are trained only on individual graphs. We believe that this work opens a new research direction in joint representation learning over multiple knowledge graphs.

Many data driven organizations such as Google and Microsoft take the approach of constructing a unified super-graph by integrating data from multiple sources. Such unification has shown to significantly help in various applications, such as search, question answering, and personal assistance. To this end, there exists a rich body of work on linking entities and relations, and conflict resolution (e.g., knowledge fusion ~\cite{DonGabHeiHorLaoMurStrSunZha14}. Still, the problem remains challenging for large scale knowledge graphs and this paper proposes a deep learning solution that can play a vital role in this construction process. In real-world setting, we envision our method to be integrated in a large scale system that would include various other components for tasks like conflict resolution, active learning and human-in-loop learning to ensure quality of constructed super-graph. However, we point out that our method is not restricted to such use cases---one can readily apply our method to directly make inference over multiple graphs to support applications like question answering and conversations.

For future work, we would like to extend the current evaluation of our work from a two-graph setting to multiple graphs. A straightforward approach is to create a unified dataset out of more than two graphs by combining set of triplets as described in Section 2, and apply learning and inference on the unified graph without any major change in the methodology. Our inductive framework learns functions to encode contextual information and hence is graph independent. Alternatively, one can develop sophisticated approaches with iterative merging and learning over pairs of graphs until exhausting all graphs in an input collection.

\section*{Acknowledgments} We would like to give special thanks to Ben London, Tong Zhao, Arash Einolghozati, Andrew Borthwick and many others at Amazon for helpful comments and discussions. We thank the reviewers for their valuable comments and efforts towards improving our manuscript. This project was supported in part by NSF(IIS-1639792, IIS-1717916).
\newpage
\bibliography{acl2018}

\begin{thebibliography}{42}
\expandafter\ifx\csname natexlab\endcsname\relax\def\natexlab#1{#1}\fi

\bibitem[{Auer et~al.(2007)Auer, Bizer, Kobilarov, Lehmann, Cyganiak, and
  Ives}]{AueBizKobLehCygIve07}
S{\"o}ren Auer, Christian Bizer, Georgi Kobilarov, Jens Lehmann, Richard
  Cyganiak, and Zachary Ives. 2007.
\newblock {DBpedia}: A nucleus for a web of open data.
\newblock In \emph{The Semantic Web}.

\bibitem[{Bollacker et~al.(2008)Bollacker, Evans, Paritosh, Sturge, and
  Taylor}]{BolEvaParStuTay08}
Kurt Bollacker, Colin Evans, Praveen Paritosh, Tim Sturge, and Jamie Taylor.
  2008.
\newblock Freebase: a collaboratively created graph database for structuring
  human knowledge.
\newblock In \emph{SIGMOD Conference}.

\bibitem[{Bordes et~al.(2014)Bordes, Glorot, Weston, and
  Bengio}]{BorGloWesBen14}
Antoine Bordes, Xavier Glorot, Jason Weston, and Yoshua Bengio. 2014.
\newblock A semantic matching energy function for learning with
  multi-relational data.
\newblock \emph{Machine Learning}.

\bibitem[{Bordes et~al.(2013)Bordes, Usunier, Garcia-Duran, Weston, and
  Yakhnenko}]{BorUsuGarWesetal13}
Antoine Bordes, Nicolas Usunier, Alberto Garcia-Duran, Jason Weston, and Oksana
  Yakhnenko. 2013.
\newblock Translating embeddings for modeling multi-relational data.
\newblock In \emph{Advances in neural information processing systems}, pages
  2787--2795.

\bibitem[{Bordes et~al.(2011)Bordes, Weston, Collobert, and
  Bengio}]{BorWesColBen11}
Antoine Bordes, Jason Weston, Ronan Collobert, and Yoshua Bengio. 2011.
\newblock Learning structured embeddings of knowledge bases.
\newblock In \emph{AAAI}.

\bibitem[{Buneman and Staworko(2016)}]{BunSta16}
Peter Buneman and Slawek Staworko. 2016.
\newblock Rdf graph alignment with bisimulation.
\newblock \emph{Proc. VLDB Endow.}

\bibitem[{Campbell et~al.(2016)Campbell, Li, Dagli, Acevedo-Aviles, Geyer,
  Campbell, and Priebe}]{CamLiDagAceGeyCamPri16}
W.~M. Campbell, Lin Li, C.~Dagli, J.~Acevedo-Aviles, K.~Geyer, J.~P. Campbell,
  and C.~Priebe. 2016.
\newblock Cross-domain entity resolution in social media.
\newblock \emph{arXiv:1608.01386v1}.

\bibitem[{Carlson et~al.(2010)Carlson, Betteridge, Kisiel, Settles,
  Hruschka~Jr., and Mitchell}]{CarBetKisSetHruMit10}
Andrew Carlson, Justin Betteridge, Bryan Kisiel, Burr Settles, Estevam~R.
  Hruschka~Jr., and Tom~M. Mitchell. 2010.
\newblock Toward an architecture for never-ending language learning.
\newblock In \emph{Proceedings of the Twenty-Fourth Conference on Artificial
  Intelligence (AAAI 2010)}.

\bibitem[{Catherine and Cohen(2016)}]{CatCoh16}
Rose Catherine and William Cohen. 2016.
\newblock Personalized recommendations using knowledge graphs: A probabilistic
  logic programming approach.
\newblock In \emph{Proceedings of the 10th ACM Conference on Recommender
  Systems}.

\bibitem[{Chen et~al.(2017)Chen, Tian, Yang, and Zaniolo}]{CheTiaYanZan17}
Muhao Chen, Yingtao Tian, Mohan Yang, and Carlo Zaniolo. 2017.
\newblock Multilingual knowledge graph embeddings for cross-lingual knowledge
  alignment.

\bibitem[{Cui et~al.(2017)Cui, Xiao, Wang, Song, Hwang, and
  Wang}]{CuiXiaWanSonHwaWan17}
Wanyun Cui, Yanghua Xiao, Haixun Wang, Yangqiu Song, Seung-won Hwang, and Wei
  Wang. 2017.
\newblock Kbqa: Learning question answering over qa corpora and knowledge
  bases.
\newblock \emph{Proc. VLDB Endow.}

\bibitem[{Dalton et~al.(2014)Dalton, Dietz, and Allan}]{DalDieAll14}
Jeffrey Dalton, Laura Dietz, and James Allan. 2014.
\newblock Entity query feature expansion using knowledge base links.
\newblock In \emph{Proceedings of the 37th International ACM SIGIR Conference
  on Research \&\#38; Development in Information Retrieval}.

\bibitem[{Dong et~al.(2014)Dong, Gabrilovich, Heitz, Horn, Lao, Murphy,
  Strohmann, Sun, and Zhang}]{DonGabHeiHorLaoMurStrSunZha14}
Xin Dong, Evgeniy Gabrilovich, Geremy Heitz, Wilko Horn, Ni~Lao, Kevin Murphy,
  Thomas Strohmann, Shaohua Sun, and Wei Zhang. 2014.
\newblock Knowledge vault: A web-scale approach to probabilistic knowledge
  fusion.
\newblock In \emph{Proceedings of the 20th ACM SIGKDD International Conference
  on Knowledge Discovery and Data Mining}, pages 601--610.

\bibitem[{Dredze et~al.(2010)Dredze, McNamee, Rao, Gerber, and
  Finin}]{DreMcNRaoGerFin10}
Mark Dredze, Paul McNamee, Delip Rao, Adam Gerber, and Tim Finin. 2010.
\newblock Entity disambiguation for knowledge base population.
\newblock In \emph{Proceedings of the 23rd International Conference on
  Computational Linguistics}.

\bibitem[{Fang et~al.(2016)Fang, Zhang, Wang, Chen, and Li}]{FanZhaWanCheLi16}
Wei Fang, Jianwen Zhang, Dilin Wang, Zheng Chen, and Ming Li. 2016.
\newblock Entity disambiguation by knowledge and text jointly embedding.
\newblock In \emph{CoNLL}.

\bibitem[{Feng et~al.(2016)Feng, Huang, Yang, and Zhu}]{FenHuaYanZhu16}
Jun Feng, Minlie Huang, Yang Yang, and Xiaoyan Zhu. 2016.
\newblock Gake: Graph aware knowledge embedding.
\newblock In \emph{COLING}.

\bibitem[{Gabrilovich and Markovitch(2009)}]{GabMar09}
Evgeniy Gabrilovich and Shaul Markovitch. 2009.
\newblock Wikipedia-based semantic interpretation for natural language
  processing.
\newblock \emph{J. Artif. Int. Res.}

\bibitem[{Globerson et~al.(2016)Globerson, Lazic, Chakrabarti, Subramanya,
  Ringaard, and Pereira}]{GloLazChaSubRinPer16}
Amir Globerson, Nevena Lazic, Soumen Chakrabarti, Amarnag Subramanya, Michael
  Ringaard, and Fernando Pereira. 2016.
\newblock Collective entity resolution with multi-focal attention.
\newblock In \emph{Proceedings of the 54th Annual Meeting of the Association
  for Computational Linguistics}.

\bibitem[{Hamilton et~al.(2017)Hamilton, Ying, and Leskovec}]{HamYinLes17}
William~L. Hamilton, Rex Ying, and Jure Leskovec. 2017.
\newblock Representation learning on graphs: Methods and applications.
\newblock \emph{arXiv:1709.05584}.

\bibitem[{He et~al.(2013)He, Liu, Li, Zhou, Zhang, and
  Wang}]{HeLiuLiZhoZhaWan13}
Zhengyan He, Shujie Liu, Mu~Li, Ming Zhou, Longkai Zhang, and Houfeng Wang.
  2013.
\newblock Learning entity representation for entity disambiguation.
\newblock In \emph{Proceedings of the 51st Annual Meeting of the Association
  for Computational Linguistics}.

\bibitem[{Huang et~al.(2015)Huang, Heck, and Ji}]{HuaHecJi15}
Hongzhao Huang, Larry Heck, and Heng Ji. 2015.
\newblock Leveraging deep neural networks and knowledge graphs for entity
  disambiguation.
\newblock \emph{arXiv:1504.07678v1}.

\bibitem[{Kadlec et~al.(2017)Kadlec, Bajgar, and Kleindienst}]{KadBajKle17}
Rudolph Kadlec, Ondrej Bajgar, and Jan Kleindienst. 2017.
\newblock Knowledge base completion: Baselines strike back.
\newblock In \emph{Proceedings of the 2nd Workshop on Representation Learning
  for NLP}.

\bibitem[{Koutra et~al.(2013)Koutra, Tong, and Lubensky}]{KouTonLub13}
Danai Koutra, HangHang Tong, and David Lubensky. 2013.
\newblock Big-align: Fast bipartite graph alignment.
\newblock In \emph{2013 IEEE 13th International Conference on Data Mining}.

\bibitem[{Le and Mikolov(2014)}]{LeMik14}
Quoc Le and Tomas Mikolov. 2014.
\newblock Distributed representations of sentences and documents.
\newblock In \emph{Proceedings of the 31st International Conference on Machine
  Learning}.

\bibitem[{Lin et~al.(2015)Lin, Liu, Sun, and Zhu}]{LinLiuSunZhu15}
Yankai Lin, Zhiyuan Liu, Maosong Sun, and Xuan Zhu. 2015.
\newblock Learning entity and relation embeddings for knowledge graph
  completion.
\newblock \emph{AAAI Conference on Artificial Intelligence}.

\bibitem[{Liu et~al.(2017)Liu, Wu, and Yang}]{LiuWuYan17}
Hanxiao Liu, Yuexin Wu, and Yimin Yang. 2017.
\newblock Analogical inference for multi-relatinal embeddings.
\newblock In \emph{Proceedings of the 34th International Conference on Machine
  Learning}.

\bibitem[{Liu and Yang(2016)}]{LiuYan16}
Hanxiao Liu and Yimin Yang. 2016.
\newblock Cross-graph learning of multi-relational associations.
\newblock In \emph{Proceedings of the 33rd International Conference on Machine
  Learning}.

\bibitem[{Liu et~al.(2016)Liu, Jiang, Evdokimov, Ling, Zhu, Wei, and
  Hu}]{LiuJiaEvdLinZhuXiaWeiHu16}
Quan Liu, Hui Jiang, Andrew Evdokimov, Zhen-Hua Ling, Xiaodan Zhu, Si~Wei, and
  Yu~Hu. 2016.
\newblock Probabilistic reasoning via deep learning: Neural association models.
\newblock \emph{arXiv:1603.07704v2}.

\bibitem[{Nickel et~al.(2016{\natexlab{a}})Nickel, Murphy, Tresp, and
  Gabrilovich}]{NicMurTreGab16}
Maximilian Nickel, Kevin Murphy, Volker Tresp, and Evgeniy Gabrilovich.
  2016{\natexlab{a}}.
\newblock A review of relational machine learning for knowledge graphs.
\newblock \emph{Proceedings of the IEEE}.

\bibitem[{Nickel et~al.(2016{\natexlab{b}})Nickel, Rosasco, and
  Poggio}]{NicRosPog16}
Maximilian Nickel, Lorenzo Rosasco, and Tomaso Poggio. 2016{\natexlab{b}}.
\newblock Holographic embeddings of knowledge graphs.

\bibitem[{Nickel et~al.(2011)Nickel, Tresp, and Kriegel}]{NicTreKri11}
Maximilian Nickel, Volker Tresp, and Hans-Peter Kriegel. 2011.
\newblock A three-way model for collective learning on multi-relational data.
\newblock In \emph{Proceedings of the 28th International Conference on Machine
  Learning (ICML-11)}, pages 809--816.

\bibitem[{Pershina et~al.(2015)Pershina, Yakout, and Chakrabarti}]{PerYakCha15}
Maria Pershina, Mohamed Yakout, and Kaushik Chakrabarti. 2015.
\newblock Holistic entity matching across knowledge graphs.
\newblock In \emph{2015 IEEE International Conference on Big Data (Big Data)}.

\bibitem[{Pujara and Getoor(2016)}]{PujGet16}
Jay Pujara and Lise Getoor. 2016.
\newblock Generic statistical relational entity resolution in knowledge graphs.
\newblock In \emph{Sixth International Workshop on Statistical Relational AI}.

\bibitem[{Rossi et~al.(2017)Rossi, Zhou, and Ahmed}]{RosZhoAhm17}
Ryan~A. Rossi, Rong Zhou, and Nesreen~K. Ahmed. 2017.
\newblock Deep feature learning for graphs.
\newblock \emph{arXiv:1704.08829}.

\bibitem[{Socher et~al.(2013)Socher, Chen, Manning, and Ng}]{SocCheManNg13}
Richard Socher, Danqi Chen, Christopher~D Manning, and Andrew Ng. 2013.
\newblock Reasoning with neural tensor networks for knowledge base completion.
\newblock In \emph{Advances in Neural Information Processing Systems}, pages
  926--934.

\bibitem[{Suchanek et~al.(2007)Suchanek, Kasneci, and Weikum}]{SucKasWei07}
Fabian~M. Suchanek, Gjergji Kasneci, and Gerhard Weikum. 2007.
\newblock Yago: A core of semantic knowledge.
\newblock In \emph{Proceedings of the 16th International Conference on World
  Wide Web}.

\bibitem[{Toutanova and Chen(2015)}]{TouChe15}
Kristina Toutanova and Danqi Chen. 2015.
\newblock Observed versus latent features for knowledge base and text
  inference.
\newblock In \emph{ACL}.

\bibitem[{Toutanova et~al.(2016)Toutanova, Lin, Yih, Poon, and
  Quirk}]{TouLinYihPooetal16}
Kristina Toutanova, Xi~Victoria Lin, Wen-tau Yih, Hoifung Poon, and Chris
  Quirk. 2016.
\newblock Compositional learning of embeddings for relation paths in knowledge
  bases and text.
\newblock In \emph{Proceedings of the 54th Annual Meeting of the Association
  for Computational Linguistics}, volume~1, pages 1434--1444.

\bibitem[{Trouillon et~al.(2016)Trouillon, Welbl, Riedel, Gaussier, and
  Bouchard}]{TroWelRieGauBou16}
Theo Trouillon, Johannes Welbl, Sebastian Riedel, Eric Gaussier, and Guillaume
  Bouchard. 2016.
\newblock Complex embeddings for simple link prediction.
\newblock In \emph{Proceedings of the 33rd International Conference on Machine
  Learning}.

\bibitem[{Wang et~al.(2014)Wang, Zhang, Feng, and Chen}]{WanZhaFenChe14}
Zhen Wang, Jianwen Zhang, Jianlin Feng, and Zheng Chen. 2014.
\newblock Knowledge graph embedding by translating on hyperplanes.

\bibitem[{Xiao et~al.(2016)Xiao, Huang, and Zhu}]{XiaHuaZhu16}
Han Xiao, Minlie Huang, and Xiaoyan Zhu. 2016.
\newblock Transg: A generative model for knowledge graph embedding.
\newblock In \emph{Proceedings of the 54th Annual Meeting of the Association
  for Computational Linguistics}.

\bibitem[{Yang et~al.(2015)Yang, Yih, He, Gao, and Deng}]{YanYihHeGaoDen15}
Bishan Yang, Wen-tau Yih, Xiaodong He, Jianfeng Gao, and Li~Deng. 2015.
\newblock Embedding entities and relations for learning and inference in
  knowledge bases.
\newblock \emph{arXiv:1412.6575}.

\end{thebibliography}
\bibliographystyle{acl_natbib}
\newpage
\onecolumn
\appendix
\section*{Appendix for }

\section{Discussion and Insights on Entity Linkage Task}
Entity linkage task is novel in the space of multi-graph learning and yet has not been tackled by any existing relational learning approaches. Hence we analyze our performance on the task in more detail here. We acknowledge that baseline methods are not tailored to the task of entity linkage and hence their low performance is natural. But we observe that our model performs well even in the unsupervised scenario where essentially the linkage loss function is switched off and our model becomes a relational learning baseline. We believe that the inductive ability of our model and shared parameterization helps to capture knowledge across graphs and allows for better linkage performance. This outcome demonstrates the merit in multi-graph learning for different inference tasks. Having said that, we admit that our results are far from comparable to state-of-the-art linkage results (Das et al., 2017) and much work needs to be done to advance representation and relational learning methods to support effective entity linkage. But we note that our model works for multiple types of entities in a very heterogeneous environment with some promising results which serves as an evidence to pursue this direction for entity linkage task.
\\\\ 
We now discuss several use-case scenarios where our model did not perform well to gain insights on what further steps can be pursued to improve over this initial model:\\\\
{\bf Han Solo with many attributes (False-negative example).} Han Solo is a fictional character in Star Wars and appears in both D-IMDB and D-FB records. We have a positive label for this sample but we do not predict it correctly. Our model combines multiple components to effectively learn across graphs. Hence we investigated all the components to check for the failures. One observation we have is the mismatch in the amount of attributes across the two datasets. Further, this is compounded by multi-value attributes. As described, we use paragraph2vec like model to learn attribute embeddings where for each attribute, we aggregate over all its values. This seems to be computing embeddings that are very noisy. As we have seen attributes are affecting the final result with high impact and hence learning very noisy attributes is not helping. Further, the mismatch in number of types is also an issue. Even after filtering the types, the difference is pretty large. Types are also included as attributes and they contribute context to relation embeddings. We believe that the skew in type difference is making the model learn bad embeddings. Specifically this happens in cases where lot of information is available like Han Solo as it lead to the scenario of abundant noisy data. With our investigation, we believe that contextual embeddings need further sophistication to handle such scenarios. Further, as we already learn relation, type and attribute embeddings in addition to entity embeddings, aligning relations, types and attributes as integral task could also be an important future direction.
\\\\
{\bf Alfred Pennyworth is never the subject of matter (False-negative example).} In this case, we observe a new pattern which was found in many other examples. While there are many triples available for this character in D-IMDB,  very few triplets are available in D-FB. This skew in availability of data hampers the learning of deep network which ends up learning very different embeddings for two realizations. Further, we observe another patter where Alfred Pennyworth appears only as an object in all those few triplets of D-FB while it appears as both subject and object in D-IMDB. Accounting for asymmetric relationships in an explicit manner may become helpful for this scenario.
\\\\
{\bf Thomas Wayne is Martha Wayne! (False-positive example).} This is the case of abundance of similar contextual information as our model predicts Thomas Wayne and Martha Wayne to be same entity. Both the characters share a lot of context and hence many triples and attributes, neighborhood etc. are similar for of them eventually learning very similar embeddings. Further as we have seen before, neighborhood has shown to be a weak context which seems to hamper the learning in this case. Finally, the key insight here is to be able to attend to the very few discriminative features for the entities in both datasets (e.g. male vs female) and hence a more sophisticated attention mechanism would help.\\\\
In addition to the above specific use cases, we would like to discuss insights on following general  concepts that naturally occur when learning over multiple graphs:
\begin{itemize}
\item {\bf Entity Overlap Across Graphs.} In terms of overlap, one needs to distinguish between *real* and *known* overlap between entities. For the known overlap between entities, we use that knowledge for linkage loss function $L_{lab}$. But our method does not need to assume either types of overlap. In case there is no real overlap, the model will learn embeddings as if they were on two separate graphs and hence will only provide marginal (if any) improvement over state-of-art embedding methods for single graphs. If there is real overlap but no known overlap (i.e., no linked entity labels), the only change is that Equation (13) will ignore the term $(1-b) \cdot L_{lab}$. Table 3 shows that in this case (corresponding to AUPRC (Unsupervised)), we are still able to learn similar embeddings for graph entities corresponding to the same real-world entity.

\item {\bf Disproportionate Evidence for entities across graphs.} While higher proportion of occurrences help to provide more evidence for training an entity embedding, the overall quality of embedding will also be affected by all other contexts and hence we expect to have varied entity-specific behavior when they occur in different proportions across two graphs

\item {\bf Ambiguity vs. Accuracy.} The effect of ambiguity on accuracy is dependent on the type of semantic differences. For example, it is observed that similar entities with major difference in attributes across graphs hurts the accuracy while the impact is not so prominent for similar entities when only their neighborhood is different.
\end{itemize}
\section{Implementation Details}
\subsection{Additional Dataset Details}
We perform light pre-processing on the dataset to remove self-loops from triples, clean the attributes to remove garbage characters and collapse CVT (Compound Value Types) entities into single triplets. Further we observe that there is big skew in the number of types between D-IMDB and D-FB. D-FB contains many non-informative type information  such as $\#base.*$. We remove all such non-informative types from both datasets which retains 41 types in D-IMDB and 324 types in D-FB.  This filtering does not reduce the number of entities or triples by significant number (less than 1000 entities filtered) 

For comparing at scale with baselines, we further reduce dataset using similar techniques adopted in producing widely accepted FB-15K or FB-237K. Specifically, we filter relational triples such that both entities in a triple contained in our dataset must appear in more than $k$ triples. We use $k=50$ for D-FB and $k=100$ for D-IMDB as D-IMDB has orders of magnitude more triples compared to D-FB in our curated datasets. We still maintain the overall ratio of the number of triples between the two datasets. 

{\bf Positive and Negative Labels.} We obtain 500662 positive labels using the existing links between the two datasets. Note that any entity can have only one positive label. We also generate 20 negative labels for each entity using the following method: (i) randomly select 10 entities from the other graph such that both entities belong to the same type and there exist no positive label between entities (ii) randomly select 10 entities from the other graph such that both entities belong to different types. 

\subsection{Training Configurations}
We performed hyper-parameter grid search to obtain the best performance of our method and finally used the following configuration to obtain the reported results:\\
-- Entity Embedding Size: 256, Relation Embedding Size=64, Attribute Embedding Size = 16, Type Embedding Size = 16, Attribute Value Embedding Size = 512. We tried multiple batch sizes with very minor difference in performance and finally used size of 2000. For hidden units per layer, we use size = 64. We used $C=50$ negative samples and $Z=20$ negative labels. The learning rate was initialized as 0.01 and then decayed over epochs. We ran our experiments for 5 epochs after which the training starts to convert as the dataset is very large. We use loss weights $b$ as 0.6 and margin as 1. Further, we use $K = 50$ random walks of length $l = 3$ for each entity   
We used a train/test split of 60\%/40\% for both the triples set and labels set. 
For baselines, we used the implementations provided by the respective authors and performed grid search for all methods according to their requirements.

\section{Contextual Information Formulations} 
Here we describe exact formulation of each context that we used in our work.\\\\
{\bf Neighborhood Context:}  Given a triplet $(e^s,r,e^o)$, the neighborhood context for an entity $e^s$ will
be all the nodes at 1-hop distance from $e^s$ other than the node $e^o$. This will capture the effect of 
other nodes in the graph surrounding $e^s$ that drives $e^s$ to participate in fact $(e^s,r,e^o)$. Concretely, we 
define the neighborhood context of $e^s$ as follows:

\begin{align}
\label{eq:neighbor}
	\mathbf{N_c}(e^s) &= \frac{1}{n_{e'}} \sum\limits_{\substack{e' \in \mathcal{N}(e^s) \\ e' \neq e^o}} \mathbf{v^{e'}}
\end{align}

where $\mathcal{N}(e^s)$ is the set of all entities in neighborhood of $e^s$ other than $e^o$. We collect the neighborhood set for each entity as a pre-processing step using a random walk method. Specifically, given a node $e$, we run $k$ rounds of random-walks of length $l$ and create the neighborhood set $\mathcal{N}(e)$ by adding all unique nodes visited across these walks.\\\\
Please note that we can also use $\max$ function in ~(\ref{eq:neighbor}) instead of sum. $\mathbf{N_c}(e^s) \in \mathbb{R}^d$ and the context
can be similarly computed for object entity.\\\\
{\bf Attribute Context.} For an entity $e^s$, the corresponding attribute context is defined as

\begin{align}
\label{eq:att}
	\mathbf{A_c}(e^s) &= \frac{1}{n_a} \sum\limits_{i=1}^{n_a} \mathbf{a_i^{e^s}}
\end{align}

where $n_a$ is the number of attributes. $\mathbf{a_i^{e^s}}$ is the embedding for attribute $i$. $\mathbf{A_c}(e^s) \in \mathbb{R}^y$.\\\\
{\bf Type Context.} We use type context mainly for relationships i.e. for a given relationship $r$, this context aims at 
capturing the effect of type of entities that have participated in this relationship. For a given triplet $(e^s, r , e^o)$, we define 
type context for relationship $r$  as:

\begin{align}
\label{eq:type}
	\mathbf{T_c}(r) &= \frac{1}{n_t^r} \sum\limits_{i=1}^{n_t^r} \mathbf{v_i^{t'}} 
\end{align}

where, $n_t^r$ is the total number of types of entities that has participated in relationship $r$ and $\mathbf{v_i^{t'}}$ is the type embedding
	that corresponds to type $t$. $\mathbf{T_c}(r) \in \mathbb{R}^q$.

\end{document}